\documentclass[sigconf]{acmart}
\usepackage{bm}
\usepackage{amsmath}
\usepackage{multirow}
\usepackage{multicol}
\usepackage{subcaption}
\usepackage{caption}
\usepackage{svg}

\usepackage{float}
\usepackage{varioref}
\usepackage{algorithm}
\usepackage[noend]{algpseudocode}
\algnewcommand{\LineComment}[1]{\State \(\triangleright\) #1}

\renewcommand\footnotetextcopyrightpermission[1]{}
\AtBeginDocument{%
  \providecommand\BibTeX{{%
    \normalfont B\kern-0.5em{\scshape i\kern-0.25em b}\kern-0.8em\TeX}}}

\setcopyright{none}

\begin{document}



\title{Learning to Infer Counterfactuals: Meta-Learning for Estimating Multiple Imbalanced Treatment Effects}

\author{Guanglin Zhou$^1$, Lina Yao$^1$, Xiwei Xu$^2$, Chen Wang$^2$, Liming Zhu$^2$}
\affiliation{%
    \institution{$^1$ University of New South Wales, Sydney, Australia, $^2$ Data61, CSIRO, Sydney, Australia}
    \country{}}
    
\email{{guanglin.zhou, lina.yao}@unsw.edu.au}
\email{{xiwei.xu, chen.wang, liming.zhu}@data61.csiro.au}

\begin{abstract}
We regularly consider answering counterfactual questions in practice, such as "Would people with diabetes take a turn for the better had they choose another medication?".  Observational studies are growing in significance in answering such questions due to their widespread accumulation and comparatively easier acquisition than Randomized Control Trials (RCTs). Recently, some works have introduced representation learning and domain adaptation into counterfactual inference. However, 
most current works focus on the setting of binary treatments. None of them considers that different treatments' sample sizes are imbalanced, especially data examples in some treatment groups are relatively limited due to inherent user preference. In this paper, we design a new algorithmic framework for counterfactual inference, which brings an idea from \textbf{Meta}-learning for Estimating \textbf{I}ndividual \textbf{T}reatment \textbf{E}ffects (MetaITE) to fill the above research gaps, especially considering multiple imbalanced treatments. Specifically, we regard data episodes among treatment groups in counterfactual inference as meta-learning tasks. We train a meta-learner from a set of source treatment groups with sufficient samples and update the model by gradient descent with limited samples in target treatment. Moreover, we introduce two complementary losses. One is the supervised loss on multiple source treatments. The other loss which aligns latent distributions among various treatment groups is proposed to reduce the discrepancy. We perform experiments on two real-world datasets to evaluate inference accuracy and generalization ability. Experimental results demonstrate that the model MetaITE matches/outperforms state-of-the-art methods.

\end{abstract}





\maketitle

\section{Introduction}
Counterfactual questions are widespread \cite{liu2020general,chernozhukov2013inference, bottou2013counterfactual, alaa2017bayesian, glass2013causal}, i.e., 
"Would people with diabetes take a turn for the better had they choose another medication". Randomized Control Trials (RCTs) are regarded as the golden standard for estimating counterfactual effects due to the randomness that the assigned treatments are not dependent on users. Nevertheless, it is expensive and time-consuming to control randomness strictly. Besides, RCTs are sometimes immoral and can not be performed in the filed of healthcare \cite{khang2015causality}. 
In contrast, observational data are pervasive and relatively easier to acquire. For instance, 
electronic health records (EHRs), which store patients' information and doctors' disease diagnosis, are typical examples of observational data. This type of data is composed of co-variables related to patients, different medications as treatments for patients, and outcomes that refer to final response effects, i.e., whether the patient recovers or not. 


\begin{figure}[!t]
\centering
\includegraphics[width=1.0\linewidth]{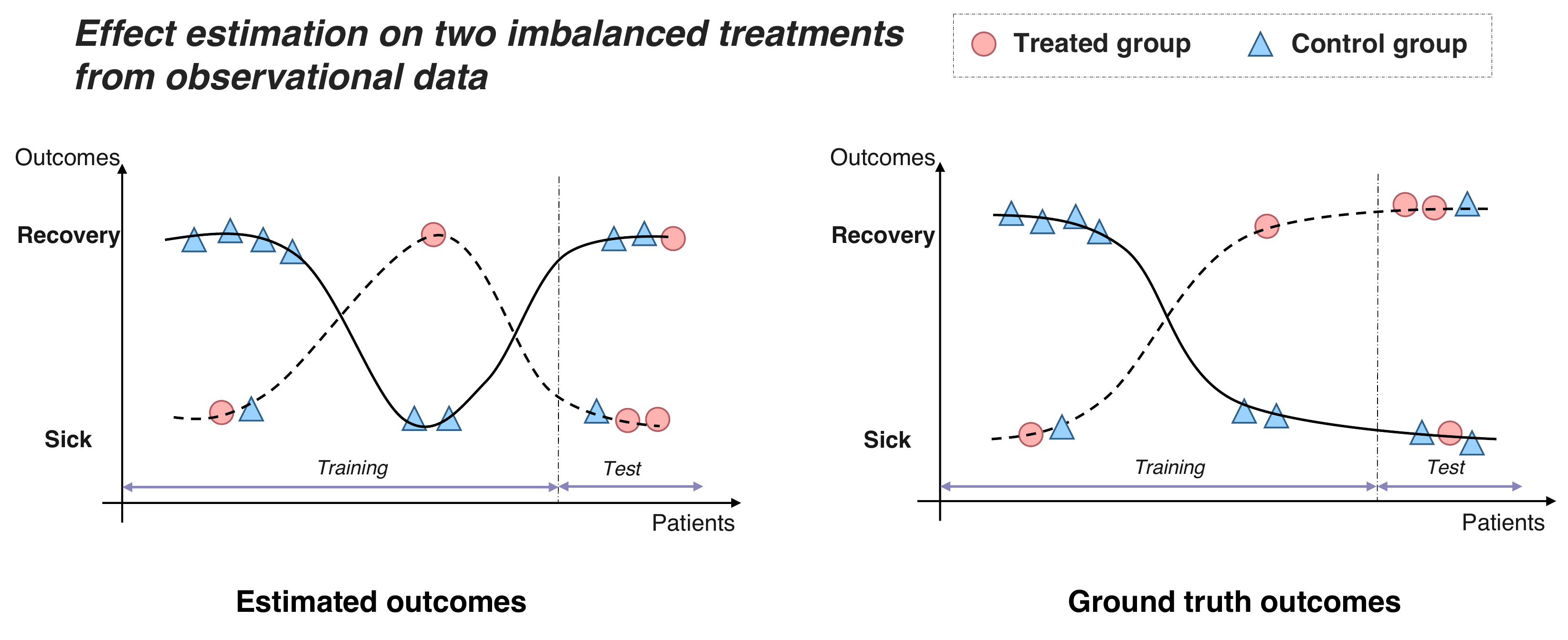}

\caption{Illustration of effect estimation on two imbalanced treatments from observational data. Consider that people with diabetes choose treatments from two drugs: the control treatment is relatively cheaper and works well, while the treated group refers to a new drug that is expensive but probably more performance-enhancing. We aim to estimate the effects for patients in test sets on these two treatments. The problem is that the control treatment seems a better choice 
(simply $\frac{4}{6} $\textgreater$ \frac{1}{2}$) 
due to the limited samples for the treated group, opposite to the ground truth in the right graph. For simplicity, we only illustrate two imbalanced treatments. }
\label{fig:logic}
\end{figure}

Recently, machine learning, especially representation learning and domain adaptation, have been demonstrated as an effective mechanism to infer counterfactual effects from observational data. The literature \cite{johansson2016learning} is the first to connect counterfactual inference with domain adaptation and introduces a regularization item that enforces similarity in the distributions between binary treatments. \cite{shalit2017estimating} extends \cite{johansson2016learning} and adopts Integral Probability Metrics (IPMs) to estimate distances between binary distributions, then formulates the model as CFR-Wass. \cite{yoon2018ganite} explores generative adversarial nets (GANs) to infer treatment effects. Considering that previous works ignore local similarity information, \cite{yao2018representation} approaches to preserve local similarity and infer effects based on deep representation learning. \cite{du2021adversarial} employs mutual information to reduce information loss, and \cite{zhou2021cycle} constructs an information loop to improve inference performance further. 

However, the methods mentioned above are all related to binary treatments except that the GANITE \cite{yoon2018ganite} is naively suitable for multiple treatments. Moreover, none of them explicitly considers the problem of imbalanced treatments. On one hand, patients face multiple choices. For example, diabetes is a disease that damages the health of many people, and it is characterized as an increase in blood sugar \cite{kooti2016role}. 
There are mainly three treatments for diabetes: drug, diet and exercise treatment. Furthermore, the drug treatment consists of oral medicine and injection medicine. Technically speaking, only modelling two treatments in many applications is not rigorous enough. 
On the other hand, imbalance is expected due to inherent self-selection in observational studies. Sample sizes among different treatment groups present significant variations and thus covariate distribution changes across treatment groups, which causes biased estimation for treatment effects.
For example, in Figure \ref{fig:logic}, samples in the control group are sufficient, while patients in the treated group are limited, probably because the new drugs are expensive and hard to afford for most patients. The imbalance causes wrong estimation for the true responses.
\cite{shalit2017estimating} also mentions imbalance issue and presents that in a real-world program called Infant Health and Development Program (IHDP), the size in the treated group is much smaller than the control group (139 treated, 608 control).
Apparently, the setting of multiple imbalanced treatments is more in line with the actual situation and research on counterfactual inference in this setting is more supportive for practical applications. 


\begin{figure*}[h!]
    \centering
    \includegraphics[width=0.8\linewidth]{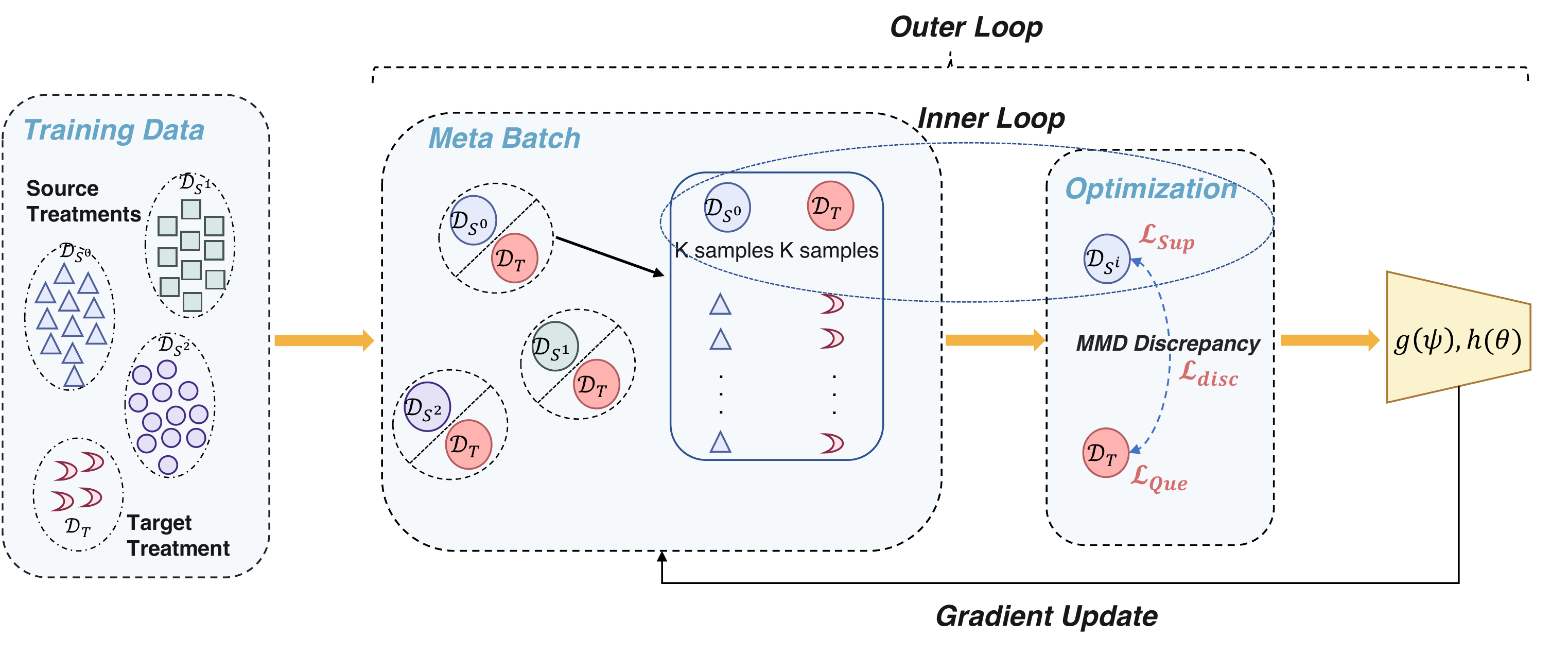}
    \caption{The overview of our model MetaITE. Data collection $\{ \mathcal{D}_{S^j}, \mathcal{D}_T \}$ from one randomly chosen source treatment group $S^j$ and target treatment group $T$. Model parameters are updated locally on the support set within the inner loop and updated globally via three losses within the outer loop. Besides, we adopt MMD discrepancy to regularize distribution disparity among treatment groups.} 
    \label{fig:MetaITE}
\end{figure*}

In order to cope with the above commonplace but ignored problems, in this paper, we propose MetaITE, a meta-learning framework for inferring individual treatment effects from observational data in the setting of multiple imbalanced treatments. Meta-learning is an efficient mechanism towards few-shot scenario and domain generalization \cite{Li2018LearningTG}. Some recent works have applied meta-learning into causality \cite{sharma2019metaci, bengio2019meta,ton2021meta}. The works \cite{bengio2019meta,ton2021meta} deal with causal directionality to distinguish the cause from effect, in contrast to counterfactual causality in our work. MetaCI is the first work that employs meta-learning for counterfactual inference \cite{sharma2019metaci}. It incorporates CFRNet \cite{johansson2016learning} into a meta-learning based reptile framework. Nevertheless, it still focuses on binary treatments and does not study the problem of imbalanced treatments. 

In this work, we focus on the setting of multiple imbalanced treatments, where samples in some groups are sufficient but limited in others. 
In particular, we cast the estimation of causal effects of multiple imbalanced treatments in a novel meta-learning framework, in which data episodes among treatment groups are regarded as meta-learning tasks. The treatment groups with plenty of samples are considered as source domains, and the treated groups with limited examples are regarded as target domains. We develop a model agnostic gradient descent framework to optimize the proposed method. We train a base meta-learner on a set of source tasks and utilize tasks in the target domain to fine-tune the learned model. We design the meta-learner with feature extractor and inference network. In addition, we make use of supervised information of the source samples. Moreover, considering distribution disparity among different treatment groups, we conceive discrepancy loss to reduce disparity via maximum mean discrepancy (MMD). As far as we know, this is the first work to apply meta-learning for estimating multiple imbalanced treatment effects, and this competitive framework is flexible and straightforward. 

The key contributions are summarized as follows. 
\begin{itemize}
    \item To the best of our knowledge, this is the first work that studies the problem of multiple imbalanced treatments in counterfactual inference.
    \item We propose a novel framework MetaITE for estimating individual treatment effects by incorporating inferring counterfactuals into the meta-learning paradigm. Moreover, we utilize the supervised information with source samples and attach discrepancy measures to reduce distribution disparity among treatment groups.
    \item Extensive experiments on two real-world datasets demonstrate MetaITE outperforms the state-of-the-arts. 
\end{itemize}


\section{Problem Setup}
\label{sec_problem}

\begin{center}
\begin{table}[h!]
\caption{Notations.}
\label{table_notation}
\begin{tabular}{|c|c|}
\hline
            $\textbf{Symbol}$      & $\textbf{Description}$                       \\ \hline
            $\mathbf{x}_{i}$                & covariates of the $i$-th unit                \\ 
            $\mathbf{t}_{i}$              & treatment assignment for $i$-th unit         \\ 
            $\mathbf{y}_i$ or $\mathbf{y}_i^{F}$     & observed/factual outcome for $i$-th unit     \\ 
            $\mathbf{y}_i^{CF}$             & counterfactual outcome for $i$-th unit       \\  
            \hline
            $\hat{\mathbf{y}}_i^{F}$ & estimated factual outcome
            for $i$-th unit       \\
            $\hat{\mathbf{y}}_i^{CF}$             & estimated counterfactual outcome
            for $i$-th unit       \\\hline
            
            $n$                    & number of units                              \\ 
            $p$                    & dimension of raw data                   \\
            $k$        & number of treatments \\ \hline
            
            $\{ {S}^j \}_{j=1}^{k-1}$ & source treatment groups \\
            $ {T} $ & target treatment group \\
            $\mathcal{D}_{S^j}, \mathcal{D}_{T}$ & sample data collection in meta-update \\

\hline            

\end{tabular}
\end{table} 
\end{center}
Let $\mathcal{X}$ denote the $p$-dimensional feature space, and $\mathcal{Y}$ represent potential outcomes. Consider obtaining a set of observational data $\{\mathbf{x}_i, \mathbf{t}_i, \mathbf{y}_i\}_{i=1}^{n}$, where $\mathbf{x}_i \in \mathcal{X} \subseteq \mathbb{R}^{p}$ represents covariates matrix related to users or patients. The $\mathbf{t}_i$ signifies the treatment assigned to the unit $i$ in fact, and $\mathbf{t}_i$ is from the set of $\mathcal{T} \in \{0, 1\}^k$, while $\mathbf{y}_i \in \mathcal{Y} \subseteq \mathbb{R}$ indicates the factual outcome we observe. 

For each unit $i$, we get covariates $\mathbf{x}_i^{(0)},\mathbf{x}_i^{(1)},...,\mathbf{x}_i^{(p-1)}$, along with the treatment $\mathbf{t}_i$ and factual outcome $\mathbf{y}_i$. Unlike prior works, we extend binary treatments to multiple settings, where $k \geq 2$. Following the potential outcome framework \cite{rubin1974estimating, rubin2005causal}, there are $k$ potential outcomes for each sample. We clarify the observed outcome as factual outcome $\mathbf{y}_i^{F}$ and unobserved outcome as counterfactual outcome $\mathbf{y}_i^{CF}$. Note that users select only one treatment for the factual outcome, and a set of outcomes for each unit associated with the remaining $k-1$ treatments are all called $\mathbf{y}_i^{CF}$. We aim to precisely estimate $\mathbf{y}_i^{F}$ and $\mathbf{y}_i^{CF}$ for each sample, given $\mathbf{x}_i$ and $\mathbf{t}_i$.  

Following the potential outcome framework \cite{rosenbaum1983central}, we make two assumptions of unconfoundedness and overlap in this work. \textit{Strong Ignorability} consists of the above two assumptions. Within the strong ignorability assumption, we can tackle the problem of approximating potential outcomes using a machine learning model $f: \mathcal{X} \times \mathcal{T} \rightarrow$ $\mathcal{Y}$.  


\textbf{Problem Definition} Now, we describe formal problem definition. Suppose that we obtain a set of observational data divided by multiple treatments, where sample sizes among different groups are various . Furthermore, considering that covariate distribution discrepancy among various treatments exists, we regard each treatment group as one domain \cite{shalit2017estimating}. We first select one treatment with limited samples as a target domain ${T}$. Then we view the remaining multiple treatments as source domains ${S}^1, \dots, {S}^{k-1}$, where some treatments have sufficient samples and others have relatively fewer samples. The number of domains equals the number of treatments.  

For simplicity, we refer to domains as assigned treatments subsequently.   
We build data collection for each domain which represented as $\{ \mathbf{X}, \mathbf{y} \}_j, j=1,2,\dots,k$. A base model is trained with data collection comprised of one source domain $\mathcal{D}_{S^j}$ and target domain $\mathcal{D}_{T}$ in meta update. Our goal is to learn the model $f$ so that it generalizes well on several domains. In that manner, via the predictive model $f$, we obtain response outcome $\mathbf{y}_i$ related to each sample $\mathbf{x}_i$ and treatment $\mathbf{t}_i$, a vector made up of factual outcome $\mathbf{y}_i^{F}$ and counterfactual outcomes  $\mathbf{y}_i^{CF}$.  


\section{The proposed method}
\label{sec_method}
On the basis of problem setup, we present a meta-learning based method for counterfactual inference in this section. In particular, we focus on multiple treatments and analyze the situation where sample sizes among treatments are imbalanced. The overview of our model is shown in Figure \ref{fig:MetaITE}. First, we introduce the base inference model and then present our MetaITE architecture for counterfactual inference.

\subsection{Base Model}
Similar to previous works, we build a prediction model by feeding covariates and treatments. The model's outcomes refer to factual and counterfactual effects when different treatments are fed into the model. The base prediction model $f$ in this work includes two parts with feature extractor and inference network, which is $f=g \circ h$. The feature extractor $g(\psi): \mathcal{X} \rightarrow \mathcal{Z}$ stands for the process that embeds raw data from input space $\mathcal{X}$ into latent representation space $\mathcal{Z}$. And $h(\theta): \mathcal{Z} \rightarrow \mathcal{Y}$ predicts the outcome with latent features.


\subsubsection{Feature Extractor}  

Due to inherent selection bias in observational studies, users who choose various treatments are in discrepancy. In other words, distributions among different treatment groups deviate from each other. Therefore, the base model aims to learn a treatment-invariant model using data samples from multiple treatment groups. Specifically, in each iteration, the base model is fed with data from one source treatment $\mathcal{D}_{S^j}$ that is selected randomly and target treatment $\mathcal{D}_{T}$, and we expect the base model generalizes well on these two domains. In such a manner, we obtain the treatment-invariant model on all domains after the outer-loop is completed.   

To a certain extent, the covariates matrices $\mathbf{X}$ can be represented as user preference as it generally involves user profiles and contextual semantics when the user makes a choice. We construct a multi-layers fully connected neural network to embed raw covariates into latent feature space, denoted as $g(\mathbf{X}; \psi)$.   
$\mathbf{Z} = g(\mathbf{X}; \psi)$, where $\mathbf{Z} \in \mathbb{R}^{d_z}$; $d_z$ is the embedding size; $g$ denotes the fully connected layers; $\psi$ is the parameter of the fully connected layers in the feature extractor.

In this section, we consider attaching discrepancy measures between two domains to enhance the latent feature transferability and improve the generalization ability of inference in multiple domains \cite{kumagai2019unsupervised, johansson2016learning, long2015learning}. The maximum mean discrepancy (MMD) is an integral probability metric that is a class of distances on probability measures \cite{gretton2012kernel}. MMD is a powerful paradigm that compares two distributions via transforming each distribution into a Reproducing Kernel Hilbert Space (PKHS). In our scenario, given two distributions from one source domain ${S^j}$ and the target domain ${T}$, we obtain corresponding latent representations $\mathbf{Z}_{{S^j}}$ and $\mathbf{Z}_{{T}}$. Then an empirical estimate of the squared MMD with two datasets is computed by 

\begin{equation}
\begin{aligned}
\label{eq_mmd}
        & \widehat{MMD}^2(\mathcal{D}_{S^j}, \mathcal{D}_{T}) 
        =
        \left \| \sum_{n=1}^{N} \frac{\mathit{k}(\cdot, \mathbf{Z}_{{S^j}})}{N} - \sum_{m=1}^{M} \frac{\mathit{k}(\cdot, \mathbf{Z}_{{T}})}{M}  \right \|_{\mathcal{H}_k}^2 \\
        &= 
        \sum_{n=1}^{N} \sum_{n=1}^{N} \frac{\mathit{k}(\mathbf{Z}_{{S^j}}, \mathbf{Z}_{{S^j}}^{'})}{N^2} 
        - 2\sum_{n=1}^{N} \sum_{m=1}^{M} \frac{\mathit{k}(\mathbf{Z}_{{S^j}}, \mathbf{Z}_{{T}})}{NM}  
        \\ &+ 
        \sum_{m=1}^{M}\sum_{m=1}^{M} \frac{\mathit{k}(\mathbf{Z}_{{T}}, \mathbf{Z}_{T}^{'})}{M^2}
\end{aligned}
\end{equation}

where $\mathit{k(\cdot, \cdot)}$ measures distance in Hilbert space and we use Gaussian kernel \cite{sriperumbudur2011universality} for distance calculation. Through MMD, we formulate $\mathcal{L}_{disc} = \widehat{MMD}^2(g(\mathbf{X}_{S^j}; \psi), g(\mathbf{X}_{T}; \psi))$ to measure the distribution discrepancy between each source domain $S^j$ and target domain $T$. By minimizing $\mathcal{L}_{disc}$, we ensure that $g(\mathbf{X};\psi)$ generates more balanced embeddings, even though $\mathbf{X}$ come from various domains. Furthermore, $g(\cdot\ ; \psi)$ can be optimized by stochastic gradient descent (SGD). 

\subsubsection{Inference Network}
Given the embedding feature vector $\mathbf{z}_i$ for each unit, we obtain the outcome prediction 
\begin{equation}
    \label{outcome}
    \hat{\mathbf{y}_i} = h(\mathbf{z}_i;\theta)
\end{equation}

where $h$ is implemented by a fully-connected layers neural network with the network parameter $\theta$, as shown in blue blocks of Figure \ref{fig:basemodel}.

For the scenario of classification, i.e., Twins dataset, a cross-entropy loss is used to optimize the inference network:
\begin{equation}
    \label{eq:cel}
    \mathcal{L}_{inf} = \frac{1}{n}\sum_{i=1}^n (\mathbf{y}_i log(\hat{\mathbf{y}}_i)+(1-\mathbf{y}_i)log(1-\hat{\mathbf{y}}_i))
\end{equation}

On the other hand, in response to the regression problem like the News dataset, the network can be optimized by minimizing the objective of the mean-squared error (MSE):

\begin{equation}
    \label{eq:mse}
    \mathcal{L}_{inf} = \frac{1}{n}\sum_{i=1}^n \left \| \mathbf{y}_i-\hat{\mathbf{y}}_i \right \|_2^2
\end{equation}

where $n$ denotes the number of samples in the training data. In the next section, we will introduce how to utilize and optimize these two losses $\mathcal{L}_{disc}$ and $\mathcal{L}_{inf}$ in the meta adaptation.

\subsection{MetaITE Architecture}
Generally, meta-learning aims to generalize well over various learning tasks from numerous domains. The learning procedure consists of three parts: 
\begin{itemize}
    \item episodic training scheme that splits available data into support sets and query sets for meta-train and meta-test respectively
    \item local update that optimizes model parameters on support sets within the inner-loop
    \item global update that updates parameters on the query sets within the outer-loop
\end{itemize}

\subsubsection{Episodic Training}
As Figure \ref{fig:MetaITE} shows, given training data from a set of domains $\mathcal{D}_S = \{\mathcal{D}_{S^j}\}_{j=1}^{k-1}$ and one target domain $\mathcal{D}_T$, we split training data into support sets for meta-train and query sets for meta-test. In particular, we uniformly sample one source domain ${S^j}$ from the source sets for meta-train and choose data collection $\mathcal{D}_T$ from the target domain for meta-test. As Algorithm \ref{alg1} illustrates, we construct a 
meta batch in each outer loop. Each batch includes two domains, one for the source and the other for the target. We randomly choose $K$ samples in selected source domain as support set $\{\mathbf{X}_i^{Sup}, \mathbf{y}_i^{Sup}\}_{i=1}^K$ for the local update, and $K$ identities from target domain as query set $\{\mathbf{X}_i^{Que}, \mathbf{y}_i^{Que} \}_{i=1}^K$ for global update, where $K$ is a hyper-parameter.

\subsubsection{Local Update}
We follow the model-agnostic meta-learning (MAML) that learns transferrable internal representations among different domains \cite{finn2017model}. The local update corresponds to the inner-loop in Algorithm \ref{alg1}. Note that we aim to ensure the base model generalizes well on multiple domains. We firstly let the model learn well on the source domain with support set $\{\mathbf{X}_i^{Sup}, \mathbf{y}_i^{Sup}\}_{i=1}^K$. The corresponding loss can be formulated as:

\begin{equation}
    \label{inner_loss}
    \mathcal{L}_{Sup} = \sum_{i=1}^{K}\mathcal{L}_{inf}(\mathbf{y}_i^{Sup}, h(g(\mathbf{X}_i^{Sup};\psi);\theta))
\end{equation}

And we use stochastic gradient descent (SGD) to optimize the parameters $\psi$ and $\theta$.

\begin{equation}
    \label{inner_grad}
    (\psi^{'}, \theta^{'}) \leftarrow (\psi, \theta)-\alpha \nabla_{\psi,\theta} \mathcal{L}_{Sup}
\end{equation}
where $\alpha$ is a fixed hyper-parameter controlling the update rate. In practice, the updated parameters $\psi^{'}$ and $\theta^{'}$ are computed using several gradient descent updates. We illustrate one gradient descent update for simplicity in Equation \ref{inner_grad}. The model parameters are trained and optimized with the performance of $f(\cdot\ ;\psi^{'}, \theta^{'})$. Correspondingly, the meta-objective can be formulated as :

\begin{equation}
    \label{inner_obj_psi}
    \mathop{min}\limits_{\psi} \mathcal{L}_{Sup}(f_{\psi^{'}}) = \mathop{min}\limits_{\psi}\sum_{\{\mathbf{X}, \mathbf{y}\} \sim p(\mathcal{D}_{S^j})} \mathcal{L}_{Sup}(\mathbf{y}, f(\mathbf{X}; \psi^{'}))
\end{equation}

\begin{equation}
    \label{inner_obj_theta}
    \mathop{min}\limits_{\theta} \mathcal{L}_{Sup}(f_{\theta^{'}}) = \mathop{min}\limits_{\theta}\sum_{\{\mathbf{X}, \mathbf{y}\} \sim p(\mathcal{D}_{S^j})} \mathcal{L}_{Sup}(\mathbf{y}, f(\mathbf{X}; \theta^{'}))
\end{equation}
where the number of sums equals to the number of gradient update in the inner loop.

\subsubsection{Global Update}
When we obtain $\psi^{'}$ and $\theta^{'}$, the loss is optimized with query set:
\begin{equation}
    \label{outer_loss}
    \mathcal{L}_{Que} = \sum_{i=1}^{K}\mathcal{L}_{inf}(\mathbf{y}_i^{Que}, h(g(\mathbf{X}_i^{Que};\psi^{'});\theta^{'}))
\end{equation}
Then, meta-optimization is performed:

\begin{equation}
    \label{outer_grad}
    (\psi, \theta) \leftarrow (\psi, \theta)-\beta \nabla_{\psi,\theta} \mathcal{L}_{Que}
\end{equation}
where $\beta$ is the meta-learning rate and set as a hyper-parameter. The global update corresponds to a meta-gradient update that involves a gradient through a gradient in the outer-loop of Algorithm \ref{alg1}.   

With the aid of local and global updates, the base model learns from source domains with sufficient data and is able to fast adapt to the target domain. In this way, we learn a model with great generalization on multiple treatments. 

\subsubsection{Meta optimization}

Although the MAML paradigm is capable of learning transferable information among multiple domains via the above three steps, in the context of counterfactual inference, we are required to provide precise outcomes related to each treatment domain. Furthermore, domain discrepancy affects inference performance. Therefore, in addition to combining the above losses on the support set and query set, we consider incorporating $\mathcal{L}_{disc}$ in Eq. (\ref{eq_mmd}) to optimize the MetaTIE jointly. The $\mathcal{L}_{disc}$ in the MetaITE criterion is formulated as:

\begin{equation}
    \label{eq_disc_meta}
    \mathcal{L}_{disc} = \widehat{MMD}^2(g(\mathbf{X}^{Sup};\psi), g(\mathbf{X}^{Que};\psi))
\end{equation}
where $\widehat{MMD}^2$ equals to Eq. (\ref{eq_mmd}). By enforcing latent distributions in the source and target domains, discrepancy among different treatment groups is reduced after the training iterations. The ultimate objective loss for meta optimization is concluded, and Adam \cite{kingma2014adam} is used. 
\begin{equation}
    \label{eq_obj_loss}
    \mathcal{L}_{obj} = \mu \mathcal{L}_{Que} + \epsilon \mathcal{L}_{Sup} + \gamma \mathcal{L}_{disc} + \left \| \omega \right \|_2
\end{equation}
The final term $\left \| \cdot \right \|_2 $ is $l_2$ regularization for model complexity. The detailed pseudo code of the training process is shown in Algorithm \ref{alg1}. Regarding the testing section, we select test data to construct query sets and sample support sets from each domain iteratively . In this way, we gain all estimated effect outcomes coordinating with each treatment, i.e., $\{ \hat{\mathbf{y}}_i^{\mathbf{t}_i=t} \}_{t=1}^{k}$. The estimate process of MetaITE is shown in Algorithm \ref{alg2}.

\begin{algorithm}
	\renewcommand{\algorithmicrequire}{\textbf{Input:}}
	\renewcommand{\algorithmicensure}{\textbf{Output:}}
	\caption{\bf{:} Training process of MetaITE}
	\label{alg1}
	\begin{algorithmic}[1]
		\State \textbf{Input:} Set of source domains $\mathcal{D}_S=\{\mathcal{D}_{S^j}\}_{j=1}^{k-1}$; Target domain $\mathcal{D}_T=\{\mathcal{D}_{T}\}$
		\State \textbf{Require:} Hyperparameters $\alpha, \beta, \mu, \epsilon, \gamma$
		\State \textbf{Output:} Feature Extractor $g(\psi)$, Inference Network $h(\theta)$
		\LineComment{learning $\psi, \theta$ from training data}
		\State randomly initialize model parameters $\psi, \theta$
		\While{Outer-loop not done}
		  \State Uniformly select one source domain ${S^j}$
		  \State Sample a batch from $\{\mathcal{D}_{S^j}, \mathcal{D}_{T}\}$
		  \While{Inner-loop not done}
		      \State $K$ samples in support set from $\mathcal{D}_{S^j}$ : $\{\mathbf{X}_{i}^{Sup}, \mathbf{y}_{i}^{Sup} \}_{i=1}^K$
  		      \State $K$ samples in the query set from $\mathcal{D}_{T}$ : $\{\mathbf{X}_{i}^{Que}, \mathbf{y}_{i}^{Que} \}_{i=1}^K$
    		  \State Compute $\mathcal{L}_{Sup}$ using Eq.(\ref{inner_loss})
    		  \State Compute adapted parameters $(\psi^{'}, \theta^{'})$ via  Eq.(\ref{inner_grad})
		  \EndWhile
		  \State \textbf{end while}
		  \State Compute $\mathcal{L}_{Que}$ with Eq.(\ref{outer_loss})
		  \State Compute discrepancy $\mathcal{L}_{disc}$ by using Eq.(\ref{eq_disc_meta})
		  \State Update $(\psi, \theta) \leftarrow (\psi, \theta)-\beta \nabla_{\psi,\theta}\mathcal{L}_{obj}$
		  
		\EndWhile
		\State \textbf{end while}
		
	\end{algorithmic}
\end{algorithm}

\section{Experiments}
\label{sec_experiment}
In this section, we evaluate model performance on two real-world datasets, Twins and News. These two datasets are specified as benchmarks in baselines, and we verify MetaITE and baselines in binary and multiple treatments. In addition to performance evaluation, we also analyze robustness on imbalance among treatment groups. 
\subsection{Datasets}

\textbf{Twins.} This dataset is created from all twins birth\footnote{http://data.nber.org/data/linked-birth-infant-death-data-vital-statistics-data.html} in the USA between 1989-1991, and we only pay attention to the twins weighing less than 2kg and without missing features \cite{louizos2017causal}. The dataset is the same as \cite{yoon2018ganite, zhou2021cycle} for binary treatments. Significantly, there are 30 covariates related to the parents, the pregnancy and the birth. We use the treatment $t=1$ as being the heavier twin, and $t=0$ is expressed as the lighter twin. The outcome is mortality after one year. The final dataset includes 11,400 pairs of twins. The selection bias is introduced as that we selectively choose one of the twins as the observation and hide the other: $t_i|x_i \sim Bern(Sigmoid(w^\mathrm{T}x_i+n))$, where $w^T\sim \mathcal{U}((-0.1,0.1)^{30\times 1})$ and $n \sim \mathcal{N}(0,01)$. Moreover, the mortality rate for the lighter twin is 17.7\%, and that in the heavier twin is 16.1\%.  Concerning multiple treatments, we follow the procedure in \cite{yoon2018ganite, louizos2017causal} and four treatments are shaped as $(1)\ t=0$: lower weight and female sex, $(2)\ t=1$: lower weight and male sex, $(3)\ t=2$: higher weight and female sex, $(4)\ t=3$: higher weight and male sex. There are 11984 samples and 50 covariates. The outcome is the same as that in binary treatments.

\noindent \textbf{News.} \cite{johansson2016learning} introduces the News dataset that simulates the opinions of a media consumer exposed to multiple news items, which is originated from the NY Times corpus$\footnote{https://archive.ics.uci.edu/ml/datasets/bag+of+words}$. The News dataset is introduced for binary treatments in \cite{johansson2016learning} and \cite{schwab2020learning} extends the original dataset to the multiple setting for estimating average dose-response curves. We assume that readers prefer to read certain content on some specific devices. For example, readers may like to read brief news items like hot topics on mobile phones but prefer to read news with professional content and longer paragraphs via computers, i.e., The Economist. Furthermore, we aim to infer the individual treatment effects of obtaining more content from some specific devices on the reader's opinions. In particular, each sample $\mathbf{x}_i$ refers to news items represented by word counts, and simulated outcome $\mathbf{y}_i \in \mathbb{R}$ represents the reader's opinions of the news. Regarding the intervention $\mathbf{t}_i$, we follow \cite{schwab2020learning} and construct multiple treatments which correspond to various devices used to view the news items, such as desktop, smartphone, newspaper, and tablet. We train an LDA topic model \cite{blei2003latent} that characterizes the topic distribution of each news item. We select 5000 samples, and 50 LDA topics are learned from the corpus. 
We define $z(\mathbf{X})$ as the topic distribution of news items, and randomly choose $k$ centroids in topic space $\{ z^0, \cdots, z^{k-1} \}$ as different devices and set $z^m$ as the mean centroid of the whole topic space. Random Gaussian distribution is derived with mean $\mu \sim \mathcal{N}(0.45, 0.15)$ and standard deviation $\sigma \sim \mathcal{N}(0.1, 0.05)$. The outcomes are firstly deduced with $\tilde{{y}} \sim \mathcal{N}(\mu, \sigma)+\epsilon$ with $\epsilon \sim \mathcal{N}(0, 0.15)$. Then for each sample, we provide $k$ potential outcomes, which is $\mathbf{y}_i^{j} \sim C(\tilde{y}\ \mathbf{D}(z(\mathbf{x}_i),z^j)+\mathbf{D}(z(\mathbf{x}_i),z^m)), j=0,\cdots,k-1$, where $C$ is a scaling factor and $\mathbf{D}$ is the Euclidean distance metric. The observed treatment is regarded as $\mathbf{t}_i|\mathbf{x}_i \sim$ softmax$(\kappa \mathbf{y}_i)$ with a treatment assignment bias $\kappa$. We set $C=50$ and $\kappa=10$ by following previous work \cite{schwab2020learning}. There are two different variants of this dataset with $k = \{2, 4\}$ viewing devices.     


We randomly separate both datasets into training and testing sets with a ratio of $80\%:20\%$.

\subsection{Baselines}
\label{sec:baselines}

We adopt the following state-of-the-art models that achieve the best performance in counterfactual inference.
\begin{itemize}
    \item {\textbf{OLS/LR$_1$}}: least square regression using the treatment as a feature. 
    \item {\textbf{OLS/LR$_2$}}: separate least square regressions for each treatment group. 
    \item {\textbf{k-NN}}: \cite{crump2008nonparametric} k-nearest neighbour method for inferring individual treatment effects.
    \item {\textbf{BNN}}:  \cite{johansson2016learning} refers to the balancing neural networks that firstly incorporates representation learning into counterfactual inference.
    \item {\textbf{CFR-Wass}\footnote{https://github.com/clinicalml/cfrnet} : \cite{shalit2017estimating} builds on and extends the work in \cite{johansson2016learning}. We add a regularization and learn a balanced representation between binary treatment groups considering selection bias in observational data. They use Wasserstein distance to measure and control discrepancy.}
    \item {\textbf{CFR-MMD} : \cite{shalit2017estimating} is a variant of CFR-Wass, which uses MMD measures instead of Wasserstein distance.}
    \item {\textbf{TARNet}: \cite{shalit2017estimating} is a variant of CFR-Wass without balancing regularization.}
    \item {\textbf{GANITE}\footnote{https://github.com/jsyoon0823/GANITE}: \cite{yoon2018ganite} uses a generative adversarial network to capture and learn counterfactual distributions. The generator generates the proxies of counterfactual outcomes and passes to infer individual treatment effects. GANITE is the only work that can be applied to multiple treatments naively.}
    \item {\textbf{SITE}\footnote{https://github.com/Osier-Yi/SITE}: \cite{yao2018representation} considers that existing estimation methods almost care about balancing the distributions of control and treated groups, but none of them notes similarity information. SITE selects triple samples and preserves local similarity information by position-dependent deep metric.}
    \item {\textbf{ABCEI}\footnote{https://github.com/octeufer/Adversarial-Balancing-based-representation-learning-for-Causal-Effect-Inference}: \cite{du2021adversarial} adopts a mutual information estimator as the regularization item to preserve helpful information for predicting counterfactual effects.}
    \item {\textbf{CBRE}\footnote{https://github.com/jameszhou-gl/CBRE}: \cite{zhou2021cycle} utilizes adversarial training to balance distributions between treated and control groups. And CBRE builds an information loop to further reduce highly predictive information loss.}
\end{itemize}

\subsection{Experimental Settings}

\subsubsection{Evaluation Metrics} 
\label{sec:metrics}
We use the Rooted Precision in Estimation of Heterogeneous Effect ($\sqrt{\epsilon_{PEHE}}$) and Mean Absolute Error on ATE ($\epsilon_{ATE}$) as performance metrics for binary treatments. The smaller the two metrics are, the better the performance is.
\begin{equation} \label{eq:pehe}
    \sqrt{\epsilon_{PEHE}} = \sqrt{\frac{1}{n}\sum_{i=1}^n ((\mathbf{y}_i^{t=1}-\mathbf{y}_i^{t=0})-(\hat{\mathbf{y}}_i^{t=1}-\hat{\mathbf{y}}_i^{t=0}))^2}
\end{equation}

\begin{equation}
    \label{eq:ate}
    \epsilon_{ATE} = \left| \frac{1}{n}\sum_{i=1}^n (\mathbf{y}_i^{t=1}-\mathbf{y}_i^{t=0})-\frac{1}{n}\sum_{i=1}^n(\hat{\mathbf{y}}_i^{t=1}-\hat{\mathbf{y}}_i^{t=0}) \right|
\end{equation}

In the context of multiple treatments, we use the root mean square error ($RMSE$) \cite{yoon2018ganite} to evaluate counterfactual effects because $\sqrt{\epsilon_{PEHE}}$ cannot be extended to the multiple treatments setting. 
\begin{equation} \label{eq:rmse}
    RMSE = \sqrt{\frac{1}{n \times \left | \mathcal{T} \right |} \sum_{i=1}^n \sum_{t_i \in \mathcal{T}} (\mathbf{y}^{t=t_i}(\mathbf{x}_i)-\hat{\mathbf{y}}^{t=t_i}(\mathbf{x}_i))^2}
\end{equation}

\begin{center}
\begin{table*}[htb]
\caption{Performance Evaluation of $\textbf{MetaITE}$ with other state-of-the-art methods on Twins and News datasets. Bold indicates the method with the best performance. The lower is the better. n.r. = not reported for reasons (null value or no convergence). }
\label{table_result}
\begin{tabular}{c|c|c|c|c|c|c}
\hline
\multirow{2}{*}{Methods}   & \multicolumn{2}{c}{Twins\_bin} &                       \multicolumn{2}{c}{News\_2} & Twins\_4 & News\_4                                    \\ \cline{2-7}
  & $\sqrt{\epsilon_{PEHE}}$ & $\epsilon_{ATE}$ & $\sqrt{\epsilon_{PEHE}}$ & $\epsilon_{ATE}$ & $RMSE$ & $RMSE$\\ \cline{1-7} 
   OLS/LR$_1$ & $0.3097\pm0.000$ & $0.0079\pm0.000$ & $16.1785 \pm 0.000$ & $16.9774\pm0.000$ & $0.2267\pm0.000$& $11.04542\pm0.000$\\
   OLS/LR$_2$ & $0.3175\pm0.000$ & ${0.0099}\pm{0.000}$ &$14.1958\pm0.000$ &$12.6928\pm0.000$ & n.r.& $16.3344\pm0.000$\\
   K-NN & $0.3101\pm0.000$ & $0.0077\pm0.000$ &$27.3473\pm0.000$ &$26.3602\pm0.000$ &$0.2359\pm0.000$ & $8.75516\pm0.000$\\
   BNN & $0.3179\pm0.000$& $0.0102\pm0.000$ &$\textbf{13.6390}\pm\textbf{0.000}$ &$12.1902\pm0.000$ & $1.4939\pm0.000$ & $9.45723\pm0.000$ \\ \cline{1-7}
   TARNet & $0.3355\pm0.000$ & $0.0101\pm0.000$ & $19.0558\pm0.000$& $19.0550 \pm 0.000$ & $0.1925\pm0.000$ & $9.2592\pm0.000$\\
   CFR-Wass & $0.3112\pm0.000$ & $0.0078\pm0.000$ & $19.2389\pm0.000$ & $19.2375\pm0.000$ & ${0.1947}\pm{0.000}$ & $9.4580\pm0.000$ \\
   CFR-MMD & $0.3355\pm0.000$& $0.0101\pm0.000$ & $19.2642\pm0.000$&$19.2628\pm0.000$ & $0.1968\pm0.000$& $10.7841\pm0.000$ \\
   GANITE & $0.3228\pm0.008$ & $0.0690\pm0.039$ & n.r. & n.r. & $0.2277\pm0.040$ & n.r. \\
   SITE & $0.3180\pm0.002$ & $0.0069\pm0.001$ & $20.5922\pm0.003$& $16.7775\pm0.000$  & $0.2004\pm0.000$ & $9.2265\pm0.000$ \\
   ABCEI & n.r. & n.r. & n.r. & n.r. & $0.2090\pm0.000$  & n.r.\\
   CBRE & $0.3302\pm0.000$ & $0.0107\pm0.000$ & $19.5432\pm0.000$&$19.542\pm0.000$ & $0.2093\pm0.000$ & $11.9447\pm 0.001$\\ \cline{1-7}
   MetaITE & $\textbf{0.3093}\pm\textbf{0.000}$ & $\textbf{0.0062}\pm\textbf{0.000}$ &${17.1726}\pm{0.359}$ & $\textbf{9.2552}\pm\textbf{0.737}$ & $\textbf{0.1921}\pm\textbf{0.000}$ & $\textbf{8.7303 }\pm\textbf{0.100}$\\

\hline

\end{tabular}
\end{table*} 
\end{center}

Considering all of these baselines except the GANITE \cite{yoon2018ganite} are tailored explicitly to binary treatments, we extend these models for multiple treatments as follows: we select one treatment as the control group and then choose one as the treated group to perform binary treatment effects estimation. Note that there are $2^k$ distinct treatment subsets for $k$ treatments, so we need to infer $2^k$ binary treatment-effects to sum up the final multiple treatment effects. All baseline models are performed with parameters as mentioned in their papers or code repositories.

\subsubsection{Parameter Setting} The parameter setting for our model is included in appendix \ref{sec:Parameter Setting}. All the models are trained by Adam optimizer, and the maximum iteration of MetaITE is set as 15000.


\subsection{Performance Evaluation}
\label{performance_evaluation}

In this section, we evaluate our model MetaITE and compare it with baselines on two benchmarks. The model is implemented with Tensorflow 1.5.1\footnote{https://www.tensorflow.org/} in Python 3.6. We perform all of the experiments on one computing device with 16-core Intel Core i7-10700 CPUs and a total of 500 GB Memory RAM. We repeat ten times and report mean and std results.   

Generally speaking, we derive datasets with two and four treatments from raw data, i.e., Twins\_bin, Twins\_4, News\_2, News\_4. We believe evaluation on these four subsets of data collection can demonstrate the inference performance and generalization ability of MetaITE for estimating multiple treatments effects.

Below we give a detailed breakdown of the results of our experiments. We first perform experiments on the Twins dataset with binary and four treatments separately. For the Twins\_bin dataset, there are $4594$ samples in the control group for $t=0$, and we configure the sample size as 80 for the treated group. As for Twins\_4 dataset, there are $6058$ and $5926$ samples in $t=0$ and $t=2$ correspondingly. The sample number of $t=1$ and sample number of $t=3$ are both 160. In this way, we simulate datasets for both binary and multiple imbalanced treatments. As shown in Table \ref{table_result}, MetaITE outperforms other state-of-the-art models for both binary and four treatments. The results in the Twins dataset, regardless of binary treatments or four treatments, show that models based on representation learning almost outperform regression-based methods, especially for multiple treatments. So we believe a learning-based method is more suitable for counterfactual inference. On the other hand, relatively simpler models like CFR-Wass perform better than complex models such as GANITE and CBRE. We consider that some treated groups have limited samples in the imbalanced setting. In this way, model complexity in these models does not match the sample size, which leads to poor generalization. The results and analyses also indicate that the MetaITE is a promising framework for estimating multiple imbalanced treatments. Note that the result of ABCEI on Twins\_bin contains null values, so we record n.r. for that.   

Then we perform experiments on the News dataset. On the News\_2 dataset, there are 1634 samples and 160 samples for $t=0$ and $t=1$. And the News\_4 contains $\{860, 80, 80, 80\}$ samples for treatment set $\{0, 1, 2, 3\}$. The test procedure is the same as that on Twins dataset. As results in Table \ref{table_result} show, MetaITE realizes the best performance on the metric $\sqrt{\epsilon_{ATE}}$ of News\_2 and $RMSE$ of News\_4, and the performance is close to the best result on the metric $\sqrt{\epsilon_{PEHE}}$. Because the code of GANITE is designed for classification, it cannot be performed for regression tasks on the News dataset so that we record n.r. for results of GANITE. 

From the results on two benchmarks, We can observe that our method MetaITE demonstrates the stable and nearly best performance in various scenarios, including classification and regression tasks. While SITE and CBRE have better performance than CFR-Wass on other regular benchmarks \cite{yao2018representation, zhou2021cycle}, they perform worse in the situation of multiple imbalanced treatments, as shown in Table \ref{table_result}. Because these two models design extra modules to preserve similarity and reduce information loss, both models have higher complexity than CFR-Wass and TARNet. We guess that is why they are not suitable for the small data setting. Meta-learning is proposed to consider a few-shot setting, and therefore it is an appropriate mechanism for multiple imbalanced treatments. Experimental results also prove that it generalizes well. Moreover, by comparing CFR-WAss with TARNet, we can see that discrepancy measure is helpful for the inference performance. We derive MMD distance as a discrepancy measure and incorporate it in MetaITE, which improves the inference performance further. We evaluate the function of the discrepancy measure in section \ref{Ablation_study}.

\subsection{Robustness Study of Unbalanced Treatments}
In order to examine the ability of our method to handle the problem of imbalanced treatments, in this section, we perform a robustness study for imbalanced treatments. Specifically, we control the imbalance among treatment groups and assess the performance of MetaITE and baselines when imbalance increases. We use Twins\_bin and News\_4 datasets for the evaluation.   

\begin{figure}[!h]
\centering
\includegraphics[width=1.0\linewidth]{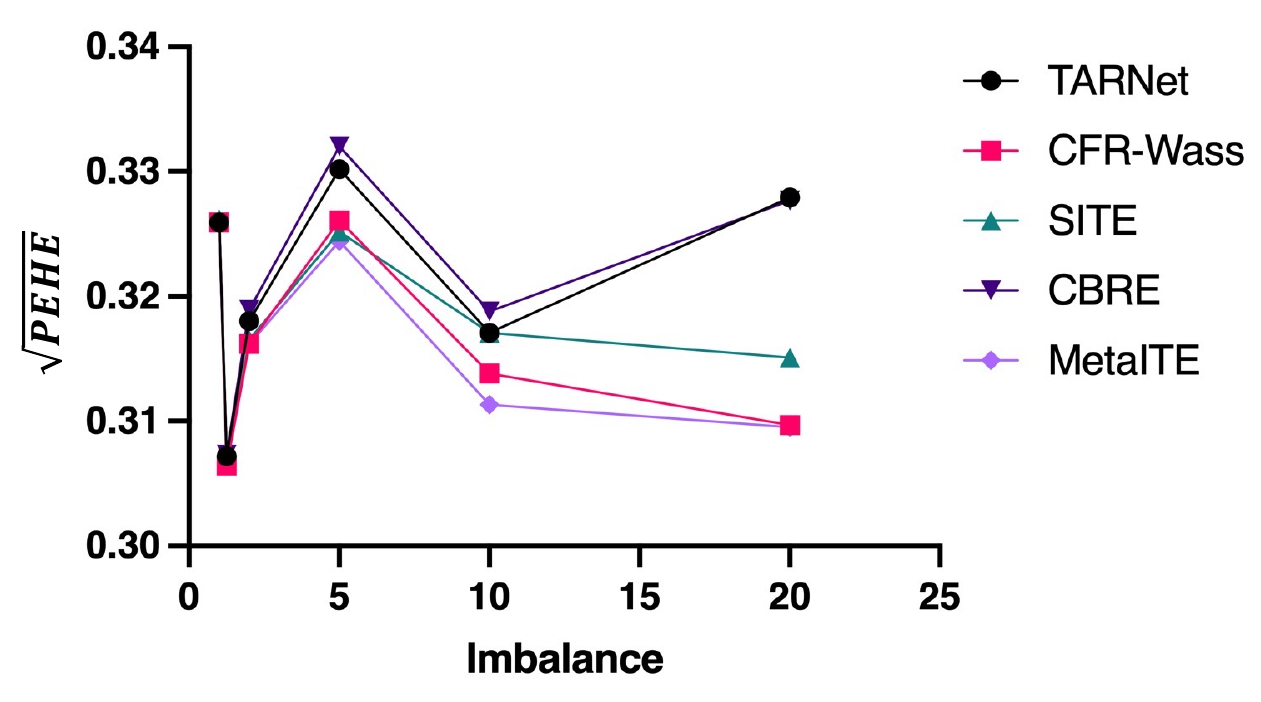}

\caption{Robustness Study of imbalanced treatments on Twins\_bin. The imbalance means that we configure this proportion of control sizes to treated samples in the experiment.}
\label{tab_twins_bin}
\end{figure}

As for the original Twins\_bin dataset, there are 4561 samples in the control group with $t=0$ and 4464 samples in the treated group with $t=1$. we keep the control group constant, and we configure the sample size of the treated group as a percentage of the initial size. We reduce the sample size from $100\%$ to $5\%$, and the smaller the percentage is, the greater the imbalance is. Correspondingly, the imbalance increases from 1 to 20. We still use $\sqrt{PEHE}$ as the performance metric for the Twins\_bin. From the results in Figure \ref{tab_twins_bin}, MetaITE outperforms four other models. As the imbalance goes larger, MetaITE has stable performance and generalizes well.




\begin{figure}[!h]
\centering
\includegraphics[width=1.0\linewidth]{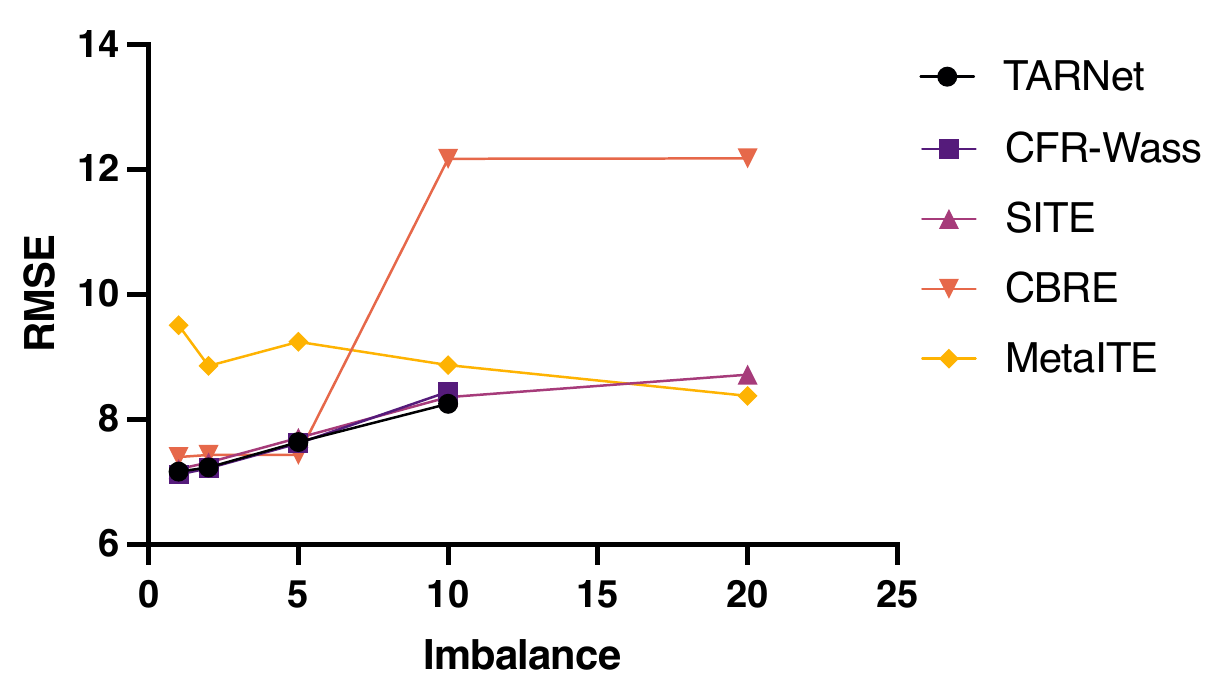}

\caption{Robustness Study of imbalanced treatments on News\_4. The imbalance means that we configure this proportion of control sizes to treated samples in the experiment. TARNet and CFR-Wass cannot solve the situation where the imbalance is greater than 10, so we record n.r. for these.}
\label{tab_news_4}
\end{figure}

Regarding to the raw News\_4 dataset, there are $\{$925, 1073, 1021, 957$\}$ samples for $t=\{$0,1,2,3$\}$. As with the process on Twins\_bin, we keep the sample size in the control group with $t=0$ and reduce the sample size from $100\%$ to $5\%$ in the other three treated groups. We adopt $RMSE$ to measure the performance of News\_4. From the results in Figure \ref{tab_news_4}, MetaITE shows a stable performance when the imbalance increases. Especially when the imbalance is large, our model still presents competitive performance compared to baselines. We find that when the imbalance decreases from $5$ to $10$, the inference performance of CBRE degrades by $39\%$. It proves that the existing complex models are not suitable for imbalanced treatments. Note that when the samples of all treatment groups are sufficient, our method is not as good as the baseline models on the News\_4 dataset. Nevertheless, treatment imbalance is more in line with real-world datasets, i.e., IHDP (139 treated, 608 control). In this work, we focus on the problem of multiple imbalanced treatments. We believe MetaITE has great potential as the applicable framework for estimating multiple imbalanced treatment effects.





\subsection{Ablation Study}
\label{Ablation_study}
The objective of MetaITE includes several loss functions. In this section, we discuss the impacts on model performance  and perform experiments on the News\_2 and News\_4 datasets. The $l_2$ regularization is set with a weight decay of 0.05. The standard meta-learning mechanism utilizes the loss on the query set to update model parameters within the outer loop and adds weight regularization. Furthermore, considering the situation of multiple imbalanced treatments, we incorporate the predictive results on source domains as extra supervision signals and combine the discrepancy measures. We configure that $\{\mu, \epsilon, \gamma\}$ are all from $0$ to $1$ with step $0.1$, and each parameter combination is evaluated 10 times. For News\_2 dataset, we first report the top three best results of $\sqrt{PEHE}$ and related hyper-parameters: $17.1726$ with $\{1.0,0.9,1.0\}$, $18.9048$ with $\{1.0,0.8,1.0\}$, $19.1451$ with $\{1.0,0.7,1.0\}$. As for News\_4, we report top three best results of RMSE and their corresponding hyper-parameters, which are 8.7303 with $\{1.0,0.0,1.0\}$, 8.9489 with $\{0.3,0.0,0.8\}$, 8.9807 with $\{0.8,0.0,1.0\}$. We see that the weight of discrepancy loss is large, which means that balancing distributions among treatment groups improves performance. On News\_4, we find that the inference loss on source treatments is not really helpful for total inference accuracy. As Eq.(\ref{outer_loss}) shows, the calculation of $\mathcal{L}_{Que}$ is related to parameters updated by $\mathcal{L}_{Sup}$. We guess these two loss functions are redundant to a certain extent on the News\_4 and on the other hand, the supervision signal of source treatments works well on the News\_2. To sum up, we obtain the optimal hyper-parameters $\{\mu,\epsilon,\gamma \}$ of $\{1.0,0.0,1.0\}$ for the News\_4 dataset and $\{1.0,0.9,1.0\}$ for the News\_2.

\section{Related work}
\label{sec_relatedwork}
\subsection{Counterfactual Inference}
Answering counterfactual questions is a fundamental problem in many applications. Researchers focus on matching methods and regression at first, such as OLS/LR and k-NN \cite{crump2008nonparametric}. The performance of inference models is greatly improved when incorporated with representation learning or neural networks. In this work, we focus on representation learning-based works. \cite{johansson2016learning} is the first framework that brings domain adaptation and representation learning for counterfactual inference. The study \cite{shalit2017estimating} learns balanced representation that enforces treated and control distributions similar via Integral Probability Metrics. \cite{yoon2018ganite} proposes GANITE that utilizes two generative adversarial networks to infer individual treatment effects. Moreover, GANITE is suitable to be applied in the setting of multiple treatments. 
Comparing previous works, Yao et al. \cite{yao2018representation} consider the local similarity information that provides meaningful regularization for inference but is ignored by most methods. 
\cite{du2021adversarial} utilizes adversarial training to balance distribution between two groups and applies mutual information metrics to reduce information loss. \cite{zhou2021cycle} extends \cite{du2021adversarial} and constructs an information loop to preserve original data properties via two supplementary losses. However, only one of them can naively infer counterfactual outcomes in the setting of multiple treatments, and none of them considers the situation of imbalanced treatments.

\subsection{Meta-Learning}
Meta-learning has been widely used in various fields, such as few-show learning, including regression and classification \cite{finn2017model} and domain generalization \cite{Li2018LearningTG}. The model-agnostic meta-learning framework is a widely adopted method called MAML \cite{finn2017model}.  In recent years, some works related to meta-learning has been utilized in causality \cite{sharma2019metaci, bengio2019meta,ton2021meta}. \cite{sharma2019metaci} incorporates CFRNet \cite{shalit2017estimating} into a meta learning-based reptile framework in order to infer counterfactual outcomes. The works \cite{bengio2019meta,ton2021meta} focus on causal discovery with meta-learning. Moreover, \cite{ton2021meta} pay attention to the small data setting and utilize a meta-learning framework, which is consistent with ours. However, they mainly deal with causal directionality to distinguish the cause from effect rather than counterfactual causality.

\section{Conclusion}
\label{sec_conclusion}
In this paper, we introduce MetaITE, a meta-learning algorithm for estimating individual treatment effects, primarily concentrating on multiple imbalanced treatments. MetaITE leverages predictive neural networks to express the feature extractor and the inference network as a base model. Moreover, it further incorporates the base model in a meta-learning framework and provides two extra regularizations: distribution discrepancy measures and the supervision of domains with high-volume samples to improve inference performance. We perform experiments on two benchmarks with real-world data, and our results empirically demonstrate that our model matches/outperforms the eleven baselines.       

In future work, we consider the existence of hidden confounders. We look forward to incorporating variational autoencoders (VAEs) into the MetaITE framework \cite{louizos2017causal}. 

\clearpage
\bibliographystyle{ACM-Reference-Format}
\bibliography{sample-base}


\begin{thebibliography}{30}


\ifx \showCODEN    \undefined \def \showCODEN     #1{\unskip}     \fi
\ifx \showDOI      \undefined \def \showDOI       #1{#1}\fi
\ifx \showISBNx    \undefined \def \showISBNx     #1{\unskip}     \fi
\ifx \showISBNxiii \undefined \def \showISBNxiii  #1{\unskip}     \fi
\ifx \showISSN     \undefined \def \showISSN      #1{\unskip}     \fi
\ifx \showLCCN     \undefined \def \showLCCN      #1{\unskip}     \fi
\ifx \shownote     \undefined \def \shownote      #1{#1}          \fi
\ifx \showarticletitle \undefined \def \showarticletitle #1{#1}   \fi
\ifx \showURL      \undefined \def \showURL       {\relax}        \fi
\providecommand\bibfield[2]{#2}
\providecommand\bibinfo[2]{#2}
\providecommand\natexlab[1]{#1}
\providecommand\showeprint[2][]{arXiv:#2}

\bibitem[\protect\citeauthoryear{Alaa and van~der Schaar}{Alaa and van~der
  Schaar}{2017}]%
        {alaa2017bayesian}
\bibfield{author}{\bibinfo{person}{Ahmed~M Alaa} {and} \bibinfo{person}{Mihaela
  van~der Schaar}.} \bibinfo{year}{2017}\natexlab{}.
\newblock \showarticletitle{Bayesian Inference of Individualized Treatment
  Effects using Multi-task Gaussian Processes}.
\newblock \bibinfo{journal}{\emph{Advances in Neural Information Processing
  Systems}}  \bibinfo{volume}{30} (\bibinfo{year}{2017}).
\newblock


\bibitem[\protect\citeauthoryear{Bengio, Deleu, Rahaman, Ke, Lachapelle,
  Bilaniuk, Goyal, and Pal}{Bengio et~al\mbox{.}}{2019}]%
        {bengio2019meta}
\bibfield{author}{\bibinfo{person}{Yoshua Bengio}, \bibinfo{person}{Tristan
  Deleu}, \bibinfo{person}{Nasim Rahaman}, \bibinfo{person}{Rosemary Ke},
  \bibinfo{person}{S{\'e}bastien Lachapelle}, \bibinfo{person}{Olexa Bilaniuk},
  \bibinfo{person}{Anirudh Goyal}, {and} \bibinfo{person}{Christopher Pal}.}
  \bibinfo{year}{2019}\natexlab{}.
\newblock \showarticletitle{A meta-transfer objective for learning to
  disentangle causal mechanisms}.
\newblock \bibinfo{journal}{\emph{arXiv preprint arXiv:1901.10912}}
  (\bibinfo{year}{2019}).
\newblock


\bibitem[\protect\citeauthoryear{Blei, Ng, and Jordan}{Blei
  et~al\mbox{.}}{2003}]%
        {blei2003latent}
\bibfield{author}{\bibinfo{person}{David~M Blei}, \bibinfo{person}{Andrew~Y
  Ng}, {and} \bibinfo{person}{Michael~I Jordan}.}
  \bibinfo{year}{2003}\natexlab{}.
\newblock \showarticletitle{Latent dirichlet allocation}.
\newblock \bibinfo{journal}{\emph{the Journal of machine Learning research}}
  \bibinfo{volume}{3} (\bibinfo{year}{2003}), \bibinfo{pages}{993--1022}.
\newblock


\bibitem[\protect\citeauthoryear{Bottou, Peters, Qui{\~n}onero-Candela,
  Charles, Chickering, Portugaly, Ray, Simard, and Snelson}{Bottou
  et~al\mbox{.}}{2013}]%
        {bottou2013counterfactual}
\bibfield{author}{\bibinfo{person}{L{\'e}on Bottou}, \bibinfo{person}{Jonas
  Peters}, \bibinfo{person}{Joaquin Qui{\~n}onero-Candela},
  \bibinfo{person}{Denis~X Charles}, \bibinfo{person}{D~Max Chickering},
  \bibinfo{person}{Elon Portugaly}, \bibinfo{person}{Dipankar Ray},
  \bibinfo{person}{Patrice Simard}, {and} \bibinfo{person}{Ed Snelson}.}
  \bibinfo{year}{2013}\natexlab{}.
\newblock \showarticletitle{Counterfactual Reasoning and Learning Systems: The
  Example of Computational Advertising.}
\newblock \bibinfo{journal}{\emph{Journal of Machine Learning Research}}
  \bibinfo{volume}{14}, \bibinfo{number}{11} (\bibinfo{year}{2013}).
\newblock


\bibitem[\protect\citeauthoryear{Chernozhukov, Fern{\'a}ndez-Val, and
  Melly}{Chernozhukov et~al\mbox{.}}{2013}]%
        {chernozhukov2013inference}
\bibfield{author}{\bibinfo{person}{Victor Chernozhukov},
  \bibinfo{person}{Iv{\'a}n Fern{\'a}ndez-Val}, {and} \bibinfo{person}{Blaise
  Melly}.} \bibinfo{year}{2013}\natexlab{}.
\newblock \showarticletitle{Inference on counterfactual distributions}.
\newblock \bibinfo{journal}{\emph{Econometrica}} \bibinfo{volume}{81},
  \bibinfo{number}{6} (\bibinfo{year}{2013}), \bibinfo{pages}{2205--2268}.
\newblock


\bibitem[\protect\citeauthoryear{Crump, Hotz, Imbens, and Mitnik}{Crump
  et~al\mbox{.}}{2008}]%
        {crump2008nonparametric}
\bibfield{author}{\bibinfo{person}{Richard~K Crump}, \bibinfo{person}{V~Joseph
  Hotz}, \bibinfo{person}{Guido~W Imbens}, {and} \bibinfo{person}{Oscar~A
  Mitnik}.} \bibinfo{year}{2008}\natexlab{}.
\newblock \showarticletitle{Nonparametric tests for treatment effect
  heterogeneity}.
\newblock \bibinfo{journal}{\emph{The Review of Economics and Statistics}}
  \bibinfo{volume}{90}, \bibinfo{number}{3} (\bibinfo{year}{2008}),
  \bibinfo{pages}{389--405}.
\newblock


\bibitem[\protect\citeauthoryear{Du, Sun, Duivesteijn, Nikolaev, and
  Pechenizkiy}{Du et~al\mbox{.}}{2021}]%
        {du2021adversarial}
\bibfield{author}{\bibinfo{person}{Xin Du}, \bibinfo{person}{Lei Sun},
  \bibinfo{person}{Wouter Duivesteijn}, \bibinfo{person}{Alexander Nikolaev},
  {and} \bibinfo{person}{Mykola Pechenizkiy}.} \bibinfo{year}{2021}\natexlab{}.
\newblock \showarticletitle{Adversarial balancing-based representation learning
  for causal effect inference with observational data}.
\newblock \bibinfo{journal}{\emph{Data Mining and Knowledge Discovery}}
  (\bibinfo{year}{2021}), \bibinfo{pages}{1--26}.
\newblock


\bibitem[\protect\citeauthoryear{Finn, Abbeel, and Levine}{Finn
  et~al\mbox{.}}{2017}]%
        {finn2017model}
\bibfield{author}{\bibinfo{person}{Chelsea Finn}, \bibinfo{person}{Pieter
  Abbeel}, {and} \bibinfo{person}{Sergey Levine}.}
  \bibinfo{year}{2017}\natexlab{}.
\newblock \showarticletitle{Model-agnostic meta-learning for fast adaptation of
  deep networks}. In \bibinfo{booktitle}{\emph{International Conference on
  Machine Learning}}. PMLR, \bibinfo{pages}{1126--1135}.
\newblock


\bibitem[\protect\citeauthoryear{Glass, Goodman, Hern{\'a}n, and Samet}{Glass
  et~al\mbox{.}}{2013}]%
        {glass2013causal}
\bibfield{author}{\bibinfo{person}{Thomas~A Glass}, \bibinfo{person}{Steven~N
  Goodman}, \bibinfo{person}{Miguel~A Hern{\'a}n}, {and}
  \bibinfo{person}{Jonathan~M Samet}.} \bibinfo{year}{2013}\natexlab{}.
\newblock \showarticletitle{Causal inference in public health}.
\newblock \bibinfo{journal}{\emph{Annual review of public health}}
  \bibinfo{volume}{34} (\bibinfo{year}{2013}), \bibinfo{pages}{61--75}.
\newblock


\bibitem[\protect\citeauthoryear{Gretton, Borgwardt, Rasch, Sch{\"o}lkopf, and
  Smola}{Gretton et~al\mbox{.}}{2012}]%
        {gretton2012kernel}
\bibfield{author}{\bibinfo{person}{Arthur Gretton}, \bibinfo{person}{Karsten~M
  Borgwardt}, \bibinfo{person}{Malte~J Rasch}, \bibinfo{person}{Bernhard
  Sch{\"o}lkopf}, {and} \bibinfo{person}{Alexander Smola}.}
  \bibinfo{year}{2012}\natexlab{}.
\newblock \showarticletitle{A kernel two-sample test}.
\newblock \bibinfo{journal}{\emph{The Journal of Machine Learning Research}}
  \bibinfo{volume}{13}, \bibinfo{number}{1} (\bibinfo{year}{2012}),
  \bibinfo{pages}{723--773}.
\newblock


\bibitem[\protect\citeauthoryear{Johansson, Shalit, and Sontag}{Johansson
  et~al\mbox{.}}{2016}]%
        {johansson2016learning}
\bibfield{author}{\bibinfo{person}{Fredrik Johansson}, \bibinfo{person}{Uri
  Shalit}, {and} \bibinfo{person}{David Sontag}.}
  \bibinfo{year}{2016}\natexlab{}.
\newblock \showarticletitle{Learning representations for counterfactual
  inference}. In \bibinfo{booktitle}{\emph{International conference on machine
  learning}}. PMLR, \bibinfo{pages}{3020--3029}.
\newblock


\bibitem[\protect\citeauthoryear{Khang}{Khang}{2015}]%
        {khang2015causality}
\bibfield{author}{\bibinfo{person}{Young-Ho Khang}.}
  \bibinfo{year}{2015}\natexlab{}.
\newblock \showarticletitle{The causality between smoking and lung cancer among
  groups and individuals: addressing issues in tobacco litigation in South
  Korea}.
\newblock \bibinfo{journal}{\emph{Epidemiology and health}}
  \bibinfo{volume}{37} (\bibinfo{year}{2015}).
\newblock


\bibitem[\protect\citeauthoryear{Kingma and Ba}{Kingma and Ba}{2014}]%
        {kingma2014adam}
\bibfield{author}{\bibinfo{person}{Diederik~P Kingma} {and}
  \bibinfo{person}{Jimmy Ba}.} \bibinfo{year}{2014}\natexlab{}.
\newblock \showarticletitle{Adam: A method for stochastic optimization}.
\newblock \bibinfo{journal}{\emph{arXiv preprint arXiv:1412.6980}}
  (\bibinfo{year}{2014}).
\newblock


\bibitem[\protect\citeauthoryear{Kooti, Farokhipour, Asadzadeh, Ashtary-Larky,
  and Asadi-Samani}{Kooti et~al\mbox{.}}{2016}]%
        {kooti2016role}
\bibfield{author}{\bibinfo{person}{Wesam Kooti}, \bibinfo{person}{Maryam
  Farokhipour}, \bibinfo{person}{Zahra Asadzadeh}, \bibinfo{person}{Damoon
  Ashtary-Larky}, {and} \bibinfo{person}{Majid Asadi-Samani}.}
  \bibinfo{year}{2016}\natexlab{}.
\newblock \showarticletitle{The role of medicinal plants in the treatment of
  diabetes: a systematic review}.
\newblock \bibinfo{journal}{\emph{Electronic physician}} \bibinfo{volume}{8},
  \bibinfo{number}{1} (\bibinfo{year}{2016}), \bibinfo{pages}{1832}.
\newblock


\bibitem[\protect\citeauthoryear{Kumagai and Iwata}{Kumagai and Iwata}{2019}]%
        {kumagai2019unsupervised}
\bibfield{author}{\bibinfo{person}{Atsutoshi Kumagai} {and}
  \bibinfo{person}{Tomoharu Iwata}.} \bibinfo{year}{2019}\natexlab{}.
\newblock \showarticletitle{Unsupervised domain adaptation by matching
  distributions based on the maximum mean discrepancy via unilateral
  transformations}. In \bibinfo{booktitle}{\emph{Proceedings of the AAAI
  Conference on Artificial Intelligence}}, Vol.~\bibinfo{volume}{33}.
  \bibinfo{pages}{4106--4113}.
\newblock


\bibitem[\protect\citeauthoryear{Li, Yang, Song, and Hospedales}{Li
  et~al\mbox{.}}{2018}]%
        {Li2018LearningTG}
\bibfield{author}{\bibinfo{person}{Da Li}, \bibinfo{person}{Yongxin Yang},
  \bibinfo{person}{Yi-Zhe Song}, {and} \bibinfo{person}{Timothy~M.
  Hospedales}.} \bibinfo{year}{2018}\natexlab{}.
\newblock \showarticletitle{Learning to Generalize: Meta-Learning for Domain
  Generalization}. In \bibinfo{booktitle}{\emph{AAAI}}.
\newblock


\bibitem[\protect\citeauthoryear{Liu, Cheng, Dong, He, Pan, and Ming}{Liu
  et~al\mbox{.}}{2020}]%
        {liu2020general}
\bibfield{author}{\bibinfo{person}{Dugang Liu}, \bibinfo{person}{Pengxiang
  Cheng}, \bibinfo{person}{Zhenhua Dong}, \bibinfo{person}{Xiuqiang He},
  \bibinfo{person}{Weike Pan}, {and} \bibinfo{person}{Zhong Ming}.}
  \bibinfo{year}{2020}\natexlab{}.
\newblock \showarticletitle{A general knowledge distillation framework for
  counterfactual recommendation via uniform data}. In
  \bibinfo{booktitle}{\emph{Proceedings of the 43rd International ACM SIGIR
  Conference on Research and Development in Information Retrieval}}.
  \bibinfo{pages}{831--840}.
\newblock


\bibitem[\protect\citeauthoryear{Long, Cao, Wang, and Jordan}{Long
  et~al\mbox{.}}{2015}]%
        {long2015learning}
\bibfield{author}{\bibinfo{person}{Mingsheng Long}, \bibinfo{person}{Yue Cao},
  \bibinfo{person}{Jianmin Wang}, {and} \bibinfo{person}{Michael Jordan}.}
  \bibinfo{year}{2015}\natexlab{}.
\newblock \showarticletitle{Learning transferable features with deep adaptation
  networks}. In \bibinfo{booktitle}{\emph{International conference on machine
  learning}}. PMLR, \bibinfo{pages}{97--105}.
\newblock


\bibitem[\protect\citeauthoryear{Louizos, Shalit, Mooij, Sontag, Zemel, and
  Welling}{Louizos et~al\mbox{.}}{2017}]%
        {louizos2017causal}
\bibfield{author}{\bibinfo{person}{Christos Louizos}, \bibinfo{person}{Uri
  Shalit}, \bibinfo{person}{Joris Mooij}, \bibinfo{person}{David Sontag},
  \bibinfo{person}{Richard Zemel}, {and} \bibinfo{person}{Max Welling}.}
  \bibinfo{year}{2017}\natexlab{}.
\newblock \showarticletitle{Causal effect inference with deep latent-variable
  models}.
\newblock \bibinfo{journal}{\emph{arXiv preprint arXiv:1705.08821}}
  (\bibinfo{year}{2017}).
\newblock


\bibitem[\protect\citeauthoryear{Rosenbaum and Rubin}{Rosenbaum and
  Rubin}{1983}]%
        {rosenbaum1983central}
\bibfield{author}{\bibinfo{person}{Paul~R Rosenbaum} {and}
  \bibinfo{person}{Donald~B Rubin}.} \bibinfo{year}{1983}\natexlab{}.
\newblock \showarticletitle{The central role of the propensity score in
  observational studies for causal effects}.
\newblock \bibinfo{journal}{\emph{Biometrika}} \bibinfo{volume}{70},
  \bibinfo{number}{1} (\bibinfo{year}{1983}), \bibinfo{pages}{41--55}.
\newblock


\bibitem[\protect\citeauthoryear{Rubin}{Rubin}{1974}]%
        {rubin1974estimating}
\bibfield{author}{\bibinfo{person}{Donald~B Rubin}.}
  \bibinfo{year}{1974}\natexlab{}.
\newblock \showarticletitle{Estimating causal effects of treatments in
  randomized and nonrandomized studies.}
\newblock \bibinfo{journal}{\emph{Journal of educational Psychology}}
  \bibinfo{volume}{66}, \bibinfo{number}{5} (\bibinfo{year}{1974}),
  \bibinfo{pages}{688}.
\newblock


\bibitem[\protect\citeauthoryear{Rubin}{Rubin}{2005}]%
        {rubin2005causal}
\bibfield{author}{\bibinfo{person}{Donald~B Rubin}.}
  \bibinfo{year}{2005}\natexlab{}.
\newblock \showarticletitle{Causal inference using potential outcomes: Design,
  modeling, decisions}.
\newblock \bibinfo{journal}{\emph{J. Amer. Statist. Assoc.}}
  \bibinfo{volume}{100}, \bibinfo{number}{469} (\bibinfo{year}{2005}),
  \bibinfo{pages}{322--331}.
\newblock


\bibitem[\protect\citeauthoryear{Schwab, Linhardt, Bauer, Buhmann, and
  Karlen}{Schwab et~al\mbox{.}}{2020}]%
        {schwab2020learning}
\bibfield{author}{\bibinfo{person}{Patrick Schwab}, \bibinfo{person}{Lorenz
  Linhardt}, \bibinfo{person}{Stefan Bauer}, \bibinfo{person}{Joachim~M
  Buhmann}, {and} \bibinfo{person}{Walter Karlen}.}
  \bibinfo{year}{2020}\natexlab{}.
\newblock \showarticletitle{Learning counterfactual representations for
  estimating individual dose-response curves}. In
  \bibinfo{booktitle}{\emph{Proceedings of the AAAI Conference on Artificial
  Intelligence}}, Vol.~\bibinfo{volume}{34}. \bibinfo{pages}{5612--5619}.
\newblock


\bibitem[\protect\citeauthoryear{Shalit, Johansson, and Sontag}{Shalit
  et~al\mbox{.}}{2017}]%
        {shalit2017estimating}
\bibfield{author}{\bibinfo{person}{Uri Shalit}, \bibinfo{person}{Fredrik~D
  Johansson}, {and} \bibinfo{person}{David Sontag}.}
  \bibinfo{year}{2017}\natexlab{}.
\newblock \showarticletitle{Estimating individual treatment effect:
  generalization bounds and algorithms}. In
  \bibinfo{booktitle}{\emph{International Conference on Machine Learning}}.
  PMLR, \bibinfo{pages}{3076--3085}.
\newblock


\bibitem[\protect\citeauthoryear{Sharma, Gupta, Prasad, Chatterjee, Vig, and
  Shroff}{Sharma et~al\mbox{.}}{2019}]%
        {sharma2019metaci}
\bibfield{author}{\bibinfo{person}{Ankit Sharma}, \bibinfo{person}{Garima
  Gupta}, \bibinfo{person}{Ranjitha Prasad}, \bibinfo{person}{Arnab
  Chatterjee}, \bibinfo{person}{Lovekesh Vig}, {and} \bibinfo{person}{Gautam
  Shroff}.} \bibinfo{year}{2019}\natexlab{}.
\newblock \showarticletitle{Metaci: Meta-learning for causal inference in a
  heterogeneous population}.
\newblock \bibinfo{journal}{\emph{arXiv preprint arXiv:1912.03960}}
  (\bibinfo{year}{2019}).
\newblock


\bibitem[\protect\citeauthoryear{Sriperumbudur, Fukumizu, and
  Lanckriet}{Sriperumbudur et~al\mbox{.}}{2011}]%
        {sriperumbudur2011universality}
\bibfield{author}{\bibinfo{person}{Bharath~K Sriperumbudur},
  \bibinfo{person}{Kenji Fukumizu}, {and} \bibinfo{person}{Gert~RG Lanckriet}.}
  \bibinfo{year}{2011}\natexlab{}.
\newblock \showarticletitle{Universality, Characteristic Kernels and RKHS
  Embedding of Measures.}
\newblock \bibinfo{journal}{\emph{Journal of Machine Learning Research}}
  \bibinfo{volume}{12}, \bibinfo{number}{7} (\bibinfo{year}{2011}).
\newblock


\bibitem[\protect\citeauthoryear{Ton, Sejdinovic, and Fukumizu}{Ton
  et~al\mbox{.}}{2021}]%
        {ton2021meta}
\bibfield{author}{\bibinfo{person}{Jean-Fran{\c{c}}ois Ton},
  \bibinfo{person}{Dino Sejdinovic}, {and} \bibinfo{person}{Kenji Fukumizu}.}
  \bibinfo{year}{2021}\natexlab{}.
\newblock \showarticletitle{Meta Learning for Causal Direction}. In
  \bibinfo{booktitle}{\emph{Proceedings of the AAAI Conference on Artificial
  Intelligence}}, Vol.~\bibinfo{volume}{35}. \bibinfo{pages}{9897--9905}.
\newblock


\bibitem[\protect\citeauthoryear{Yao, Li, Li, Huai, Gao, and Zhang}{Yao
  et~al\mbox{.}}{2018}]%
        {yao2018representation}
\bibfield{author}{\bibinfo{person}{Liuyi Yao}, \bibinfo{person}{Sheng Li},
  \bibinfo{person}{Yaliang Li}, \bibinfo{person}{Mengdi Huai},
  \bibinfo{person}{Jing Gao}, {and} \bibinfo{person}{Aidong Zhang}.}
  \bibinfo{year}{2018}\natexlab{}.
\newblock \showarticletitle{Representation learning for treatment effect
  estimation from observational data}.
\newblock \bibinfo{journal}{\emph{Advances in Neural Information Processing
  Systems}}  \bibinfo{volume}{31} (\bibinfo{year}{2018}).
\newblock


\bibitem[\protect\citeauthoryear{Yoon, Jordon, and Van Der~Schaar}{Yoon
  et~al\mbox{.}}{2018}]%
        {yoon2018ganite}
\bibfield{author}{\bibinfo{person}{Jinsung Yoon}, \bibinfo{person}{James
  Jordon}, {and} \bibinfo{person}{Mihaela Van Der~Schaar}.}
  \bibinfo{year}{2018}\natexlab{}.
\newblock \showarticletitle{GANITE: Estimation of individualized treatment
  effects using generative adversarial nets}. In
  \bibinfo{booktitle}{\emph{International Conference on Learning
  Representations}}.
\newblock


\bibitem[\protect\citeauthoryear{Zhou, Yao, Xu, Wang, and Zhu}{Zhou
  et~al\mbox{.}}{2021}]%
        {zhou2021cycle}
\bibfield{author}{\bibinfo{person}{Guanglin Zhou}, \bibinfo{person}{Lina Yao},
  \bibinfo{person}{Xiwei Xu}, \bibinfo{person}{Chen Wang}, {and}
  \bibinfo{person}{Liming Zhu}.} \bibinfo{year}{2021}\natexlab{}.
\newblock \showarticletitle{Cycle-Balanced Representation Learning For
  Counterfactual Inference}.
\newblock \bibinfo{journal}{\emph{arXiv preprint arXiv:2110.15484}}
  (\bibinfo{year}{2021}).
\newblock


\end{thebibliography}

\clearpage

\appendix
\section{APPENDIX}
\subsection{Estimate Process}
\begin{algorithm}
	\renewcommand{\algorithmicrequire}{\textbf{Input:}}
	\renewcommand{\algorithmicensure}{\textbf{Output:}}
	\caption{\bf{:} Estimate process of MetaITE}
	\label{alg2}
	\begin{algorithmic}[1]
		\State \textbf{Input:} Feature Extractor $g(\psi)$, Inference Network $h(\theta)$; Training Set and Testing Set in all domains; Hyperparameters $\alpha, \beta, \mu, \epsilon, \gamma$.
		\LineComment{estimate}
		\While{estimate not done}
		    \State construct query set $\{\mathbf{X}^{Que}, \mathbf{y}^{Que} \}$ with $K$ samples from
		    \Statex $\quad$  testing data
		    \For{$t \in \mathcal{T}$}
                \State randomly sample $K$ units from training data with 
                \Statex $\quad$ $\quad$ treatment that equals to $t$ for $\{ \mathbf{X}^{Sup}, \mathbf{y}^{Sup} \}$
                \State feed $\{ \mathbf{X}^{Sup}, \mathbf{y}^{Sup}, \mathbf{X}^{Que}, \mathbf{y}^{Que} \}$ into trained model, and 
                \Statex $\quad$ $\quad$ get prediction $\{ \hat{\mathbf{y}}_i^{\mathbf{t}_i=t} \}_{i=1}^{K}$
            \EndFor
            \textbf{End for}
            \State concatenate predictions in for-loop and obtain outcome for 
            \Statex $\quad$ each treatment $\{ \hat{\mathbf{y}}_i^{\mathbf{t}_i=t} \}_{t=1}^{k}$
		    
		\EndWhile
		\State  \textbf{end while}
		
	\end{algorithmic}
\end{algorithm}

\subsection{Parameter Setting} 
\label{sec:Parameter Setting}
There are three types of parameters: meta setup, model structure and hyper parameters.  

\begin{enumerate}
    \item Meta setup parameters: we set meta batch size as 5, which means the number of tasks sampled in each meta-update. The number of samples in the support and query sets are both set as 8. Note that the product of meta batch size and the samples number in support set or query set corresponds to the value of $K$ in Algorithm  \ref{alg1}. The number of inner gradient updates during training is 4.
    \item Model structure parameters: these parameters consist of the number of layers and the embedding size, which are related to both feature extractor and inference network. We build two-layers fully connected networks for the feature extractor with embedding size $[256, 128]$ and four-layers fully connected networks for the inference network with embedding size $[128, 128, 64, 64]$.
    \item Hyper-parameters: we mainly have five hyper-parameters $\{\alpha, \beta, \mu, \epsilon, \gamma\}$ in Algorithm \ref{alg1}. For the meta-learning rate $\beta$ and update learning rate $\alpha$ in the inner loop, we set both $\alpha$ and $\beta$ as $1e-3$. As for three coefficients of loss functions, we set the parameters $\{\mu, \epsilon, \gamma\}$ to range from $0$ to $1$ with step $0.1$ and obtain the optimal parameters.
\end{enumerate}

\subsection{Parameter Sensitivity}
\begin{figure}[!htb]
\centering

\begin{subfigure}[t]{.45\linewidth}
  \centering
  \includegraphics[width=\textwidth]{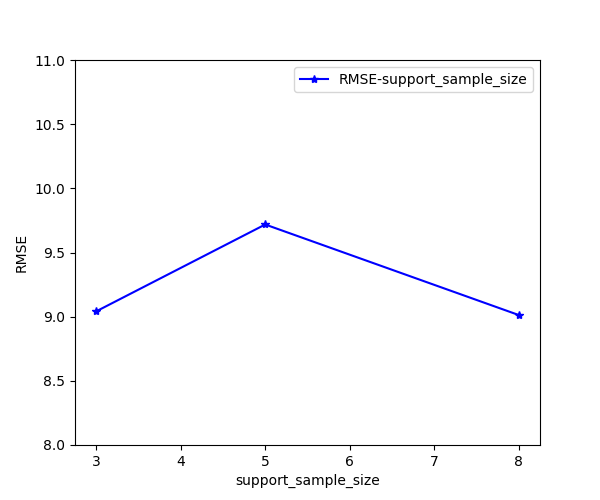}  
  \caption{Support sample size}
  \label{fig:a}
\end{subfigure}
\begin{subfigure}[t]{.45\linewidth}
  \centering
  \includegraphics[width=\textwidth]{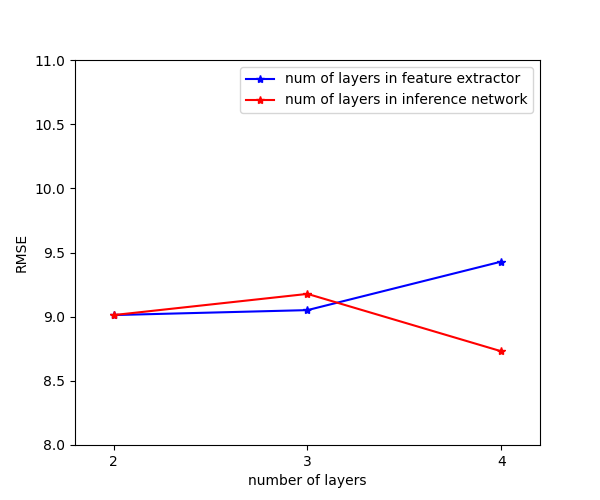}  
  \caption{Number of layers in feature extractor and inference network}
  \label{fig:b}
\end{subfigure}


\begin{subfigure}[t]{.45\linewidth}
  \centering
  \includegraphics[width=\textwidth]{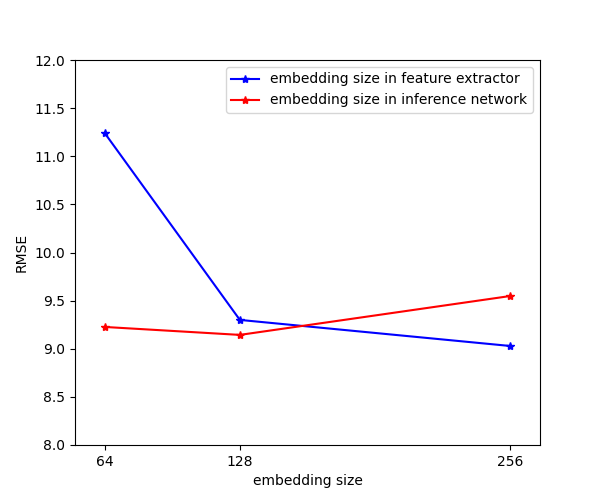}  
  \caption{Embedding size in feature extractor and inference network}
  \label{fig:c}
\end{subfigure}
\begin{subfigure}[t]{.45\linewidth}
  \centering
  \includegraphics[width=\textwidth]{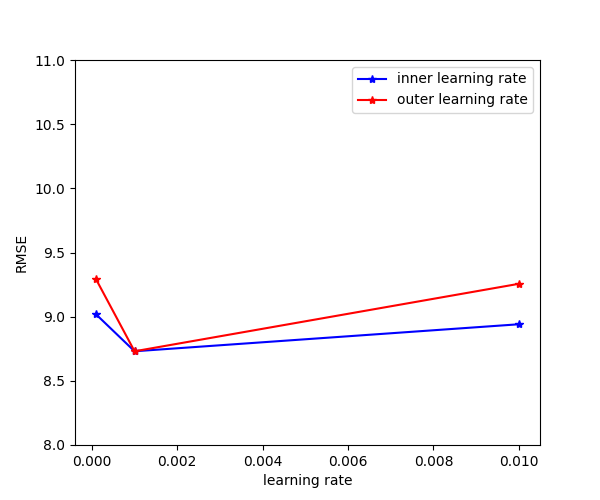}  
  \caption{Learning rate $\alpha, \beta$}
  \label{fig:d}
\end{subfigure}

\begin{subfigure}[t]{.45\linewidth}
  \centering
  \includegraphics[width=\textwidth]{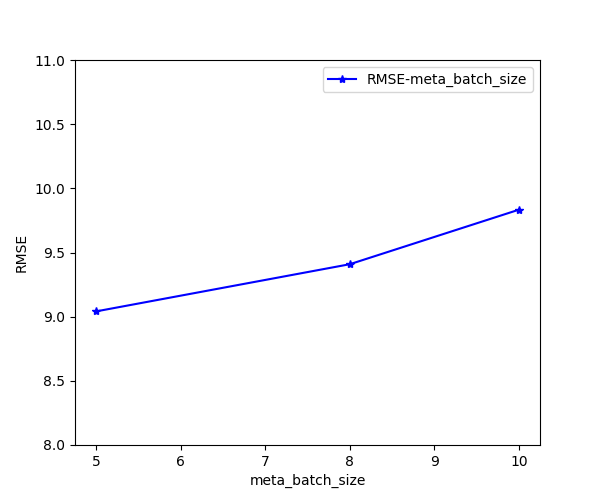}  
  \caption{Meta batch size}
  \label{fig:e}
\end{subfigure}
\begin{subfigure}[t]{.45\linewidth}
  \centering
  \includegraphics[width=\textwidth]{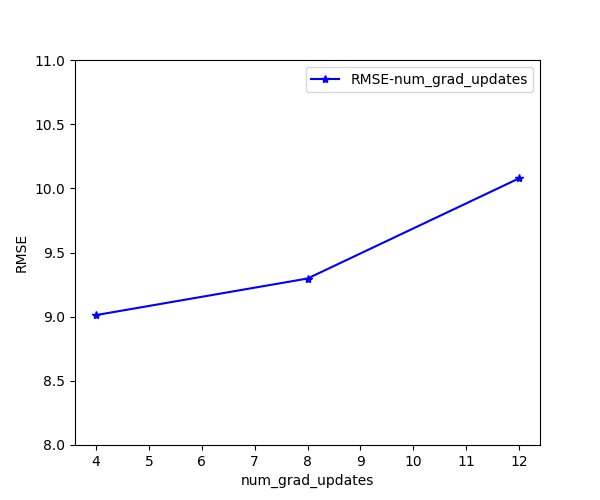}  
  \caption{Number of gradient updates}
  \label{fig:f}
\end{subfigure}

\caption{Parameter sensitivity in different settings.}
\label{fig:param_sensitivity}
\end{figure}

As introduced in the parameter setting, MetaITE contains meta setup parameters, model structure parameters and hyper-parameters. In this subsection, we evaluate all parameters except $\{\mu, \epsilon, \gamma\}$.  
Meta bach size and the number of samples in the support and the query set can influence the number of data samples fed into the local and global updates. As Figure \ref{fig:param_sensitivity} (\subref{fig:e}) shows, smaller batch size is helpful for the model performance where the size is between 5 to 10. As for the sample size in support or query set, we compare the results with size in $\{3, 5, 8\}$. In Figure \ref{fig:param_sensitivity} (\subref{fig:a}), we can find that the median value may lead to a bad result. The number of updates in the inner loop is configured from 4 to 12, and  Figure \ref{fig:param_sensitivity} (\subref{fig:f}) illustrates that more minor gradient updates provide better results, which is related to data size. According to Figure \ref{fig:param_sensitivity} (\subref{fig:b})-(\subref{fig:c}), which shows the impact of the number of layers and embedding size on model performance, we can see that shallower layers in feature extractor and deeper layers in inference network lead to better results. Moreover, the median value of embedding size  is more suitable due to the comparatively small data setting. Similarly, two learning rates $\alpha, \beta$ are chosen from $\{1e-4, 1e-3, 1e-2\}$ and median value $1e-3$ leads to better results, as seen in Figure \ref{fig:param_sensitivity} (\subref{fig:d}).

\end{document}